%% file: root.tex
\documentclass[letterpaper, 10 pt, conference]{ieeeconf}  % Comment this line out
\IEEEoverridecommandlockouts                              % This command is only
\usepackage{amsmath} % assumes amsmath package installed
\usepackage{amssymb}  % assumes amsmath package installed

\usepackage[ruled]{algorithm2e}

\usepackage{etoolbox}
\usepackage{todonotes}
\usepackage{soul}

\newcommand{\super}[1]{^{\scriptscriptstyle{\mathit{#1}\!}}}
\newcommand{\sub}[1]{_{\scriptscriptstyle{\!\mathit{#1}}}}

\newcommand{\MATRIX}[4]{%
	\ifstrempty{#2}
		{{ #1\super{#4}\sub{#3} }}
		{{ \super{#2\,}\bm{#1}\super{#4}\sub{#3} }}
}

\newcommand{\M}[3][] {\MATRIX{#2}{#1}{#3}{}}

\newcommand{\R}[2][]{\M[#1]{R}{#2}}

\undef{\v}
\newcommand{\v}[2][]{\M{\vec{#2}}{#1}}

\undef{\t}
\newcommand{\t}[1]{\v[#1]{t}}

\title{\LARGE \bf
Continuous close-range 3D object pose estimation
}

\author{Bjarne Grossmann$^*$, Francesco Rovida$^*$ and Volker Kr{\"u}ger$^{\dagger *}$% <-this % stops a space
\thanks{The research leading to these results has received funding from the European Commission's Horizon 2020 Programme under grant agreement no. ~723658 (SCALABLE).}% <-this % stops a space
\thanks{$^*$Robotics, Vision, and Machine Intelligence (RVMI) Lab at Aalborg University Copenhagen, 2450 Copenhagen, Denmark; $^\dagger$Robotics and Semantic Systems (RSS), Lund University, Lund, Sweden 
{\tt\small e-mail: bjarne|francesco@mp.aau.dk; volker.krueger@cs.lth.se}}%
}

\begin{document}

\maketitle
\thispagestyle{empty}
\pagestyle{empty}
%\renewcommand{\baselinestretch}{0.98}

%%%%%%%%%%%%%%%%%%%%%%%%%%%%%%%%%%%%%%%%%%%%%%%%%%%%%%%%%%%%%%%%%%%%%%%%%%%%%%%%
\input{abstract}
\input{introduction}
\input{approach}
% \input{implementation}

\input{evaluation}

\input{conclusions}

%%%%%%%%%%%%%%%%%%%%%%%%%%%%%%%%%%%%%%%%%%%%%%%%%%%%%%%%%%%%%%%%%%%%%%%%%%%%%%%%

 \addtolength{\textheight}{-4cm}   % This command serves to balance the column lengths

%%%%%%%%%%%%%%%%%%%%%%%%%%%%%%%%%%%%%%%%%%%%%%%%%%%%%%%%%%%%%%%%%%%%%%%%%%%%%%%%

% \section*{ACKNOWLEDGMENT}

% The preferred spelling of the word ÒacknowledgmentÓ in America is without an ÒeÓ after the ÒgÓ. Avoid the stilted expression, ÒOne of us (R. B. G.) thanks . . .Ó  Instead, try ÒR. B. G. thanksÓ. Put sponsor acknowledgments in the unnumbered footnote on the first page.

%%%%%%%%%%%%%%%%%%%%%%%%%%%%%%%%%%%%%%%%%%%%%%%%%%%%%%%%%%%%%%%%%%%%%%%%%%%%%%%%
\bibliographystyle{IEEEtran}
\bibliography{references}

\end{document}

%% file: abstract.tex
\begin{abstract}
In the context of future manufacturing lines, removing fixtures will be a fundamental step to increase the flexibility of autonomous systems in assembly and logistic operations.
Vision-based 3D pose estimation is a necessity to accurately handle objects that might not be placed at fixed positions during the robot task execution.
Industrial tasks bring multiple challenges for the robust pose estimation of objects such as difficult object properties, tight cycle times and constraints on camera views.
In particular, when interacting with objects, we have to work with close-range partial views of objects that pose a new challenge for typical view-based pose estimation methods.

In this paper, we present a 3D pose estimation method based on a gradient-ascend particle filter that integrates new observations on-the-fly to improve the pose estimate.
Thereby, we can apply this method online during task execution to save valuable cycle time.
In contrast to other view-based pose estimation methods, we model potential views in full 6-dimensional space that allows us to cope with close-range partial objects views. 
We demonstrate the approach on a real assembly task, in which the algorithm usually converges to the correct pose within 10-15 iterations with an average accuracy of less than 8mm. 
\end{abstract}

%% file: introduction.tex
\section{INTRODUCTION}
\begin{figure}[!t]
    \vspace{1ex}
    \centering
    \includegraphics[width=\columnwidth]{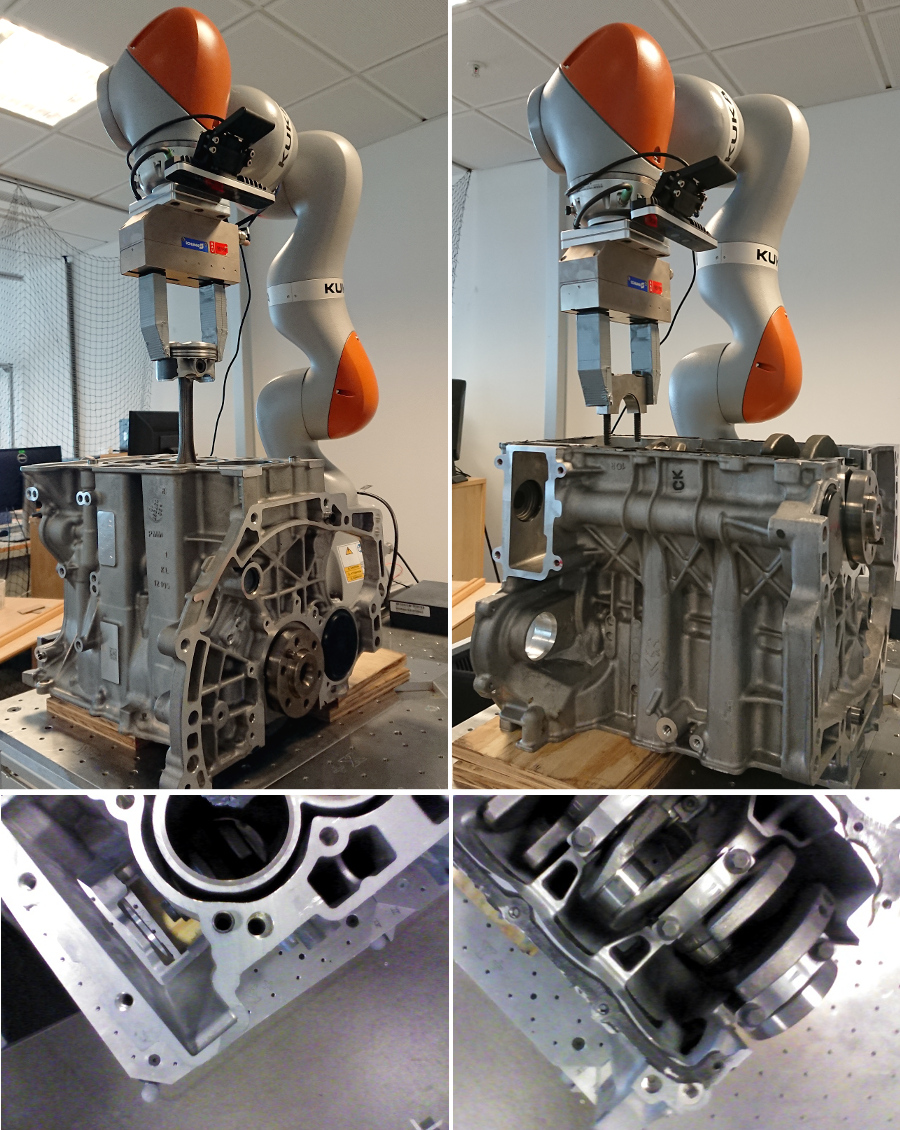}
    \caption{Robot setup for the (left) piston and (right) crankcase cap insertion with a wrist-mounted camera. During the task execution, only close-range partial views of the engine block are available posing a major challenge for the pose estimation.}
    \label{fig:robot}
    \vspace{-4ex}
\end{figure}

\noindent
With the endeavor of industrial manufacturers towards agile factories with flexible production lines, robots become more and more self-contained autonomous working units equipped with sensory devices to interact with their environment.
Workstations are not specially designed for the robot, i.e. there are no fixtures for the objects or cameras to observe the work space, and the work pieces can also be in motion on a conveyor belt or moved around by a human.
In these dynamic environments, one crucial ability of the robot to autonomously perform common manufacturing tasks such as picking, placing or assembling, is to visually estimate the accurate pose of objects, often by the means of a \mbox{RGB-D} camera that is mounted on the robot's wrist for the sake of flexibility (fig.~\ref{fig:robot}, top).
Additionally, we often have to deal with tight cycle times, and we therefore we cannot afford to have a time-consuming pose estimation procedure that e.g. includes finding better views, but we have to solve the problem during the task execution.

However, when approaching and interacting with an object, we are facing a new challenge in order to maintain a good pose estimate:
the closer we get to the object, the less we see of it.
When dealing with objects at \emph{close range}, we can therefore only use \emph{partial observations} of the object (fig.~\ref{fig:robot}, bottom) to estimate the object's pose.
Additionally, we need to continuously update the pose estimate due to the object's potential movements, which requires basic \emph{tracking capabilities} of the algorithm.

% In this scenario, a new challenge arises for the pose estimation in order to maintain a good pose estimate while approaching and interacting with object:
% The closer we get to the object, the less we see of it.
% When dealing with large objects or objects at close range, we can therefore only use \emph{partial observations} of the object (fig.~\ref{fig:robot}, bottom) to estimate the object's pose.
% Additionally, we need to continuously update the pose estimate due to the object's potential movements, which requires \emph{basic tracking capabilities} of the algorithm.

In \cite{Grossmann2015}, we showed that the pose estimation of industrial objects is due to the difficult object properties by itself already a highly challenging task.
Especially local feature-based methods \cite{Guo2016} struggle with industrial objects, because they require robust surface patches that can often not be provided by the depth sensors.
View-based methods, including descriptor-based \cite{Aldoma2012} and learning-based \cite{Xiang2017} methods, often used in tracking algorithms (tracking-by-detection) \cite{Hinterstoisser2012}, are based on manually captured or prerendered views of a CAD model.
They rephrase the pose estimation problem as a recognition or classification task by finding the best (most similar) view among the generated views with respect to the observed scene \cite{Lepetit2004}.
However, the number of rendered views is usually a trade-off between speed and robustness, thus restricting the accuracy of the pose estimate.
Additionally, the pose estimation problem can often not be solved by using one-shot pose estimation approaches due to view ambiguities \cite{Roy2004}.
Next-best-view methods \cite{Connolly1985} are trying to compensate for noise and uncertainties by finding a more informative view or integrating multiple views of a static object to improve the pose estimate, however, they are not suited for moving objects and additionally increase the cycle time of the task execution.
Integrating multiple views in a probabilistic way using Bayesian propagation over time (\cite{Riedel2016, Erkent2016}) has been recently explored, but the approach only considers the rotational space due to the high computational demands or rely on local feature descriptors.
Additionally, all view-based approaches rely on the assumption of a fully visible object at a reasonable distance.
Situation in which we have to deal with partial observations of the object from a close range have not been considered.

% However, the major issue of all these view-based methods is that they do not consider situations where we have to deal with partial observations of the object from a close range.

% In this paper, we discuss how this additional constraint is a general obstacle for all approaches relying on precomputed view-based models.
In this paper, we therefore discuss how close-range partial observations constitute an inherent problem for typical view-based pose estimation methods due to the way of how the object model is represented.
Furthermore, we provide a solution to the close-range pose estimation problem based on our previous approach in \cite{Grossmann2017}, which uses an adapted particle filter to continuously estimate the pose of a known object in almost real-time without relying on precomputed views of the object model.
Here, we extend the approach to deal with close-range observations, even when the target object is slightly moving.
We demonstrate our solution on a realistic use case: a challenging assembly task of a piston and crankcase cap (fig.~\ref{fig:robot}) that have to be inserted into a moving engine block.

% However, we did not consider close-range partial observations of the object, but assumed that the object is always fully in view.
% Here, we extend \cite{Grossmann2017} and develop a more robust structure for the particle filter that is able to deal with close-range observations and that is also able to track moving objects.
% \todo{revise}
% We demonstrate our solution on a realistic use case: a challenging assembly task of a piston and crankcase cap (see Fig.~\ref{fig:robot}) that have to be inserted into a moving engine block...
% \todo{Expand this..}

In the following section, we discuss the impact of close-range observation on the view-based pose estimation methods in detail.
Section III presents our solution to the problem and describes its implementation, whereas Section IV validates the method by means of an experimental evaluation.
In section V, we summarize our findings and outline the limitation of the proposed method.

% In the following section, we provide a mathematical background for our approach on the sequential Bayesian pose estimation.
% Section III discusses the implementation of the particle filter with its adaptions and tweaks, whereas Section IV validates the method by means of an experimental evaluation.
% Finally we conclude the paper in section V.
% \todo{Update this..}

\begin{figure}[!t]
    \vspace{1ex}
    \centering
    \includegraphics[width=0.5\columnwidth]{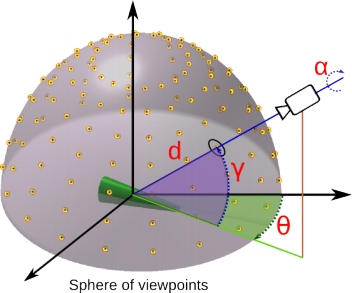}
    \caption[]{View generation from a 3D model \cite{Erkent2016}: An synthetic view is generated from each point on a sphere. However, this model covers only 4-dimensions of the object's pose.}
    \label{fig:view_generation}
    \vspace{-3ex}
\end{figure}

\section{The problem of close-range observations}
\noindent
View-based pose estimation methods usually rely on a 3D representation of an object, manually captured beforehand or often given as a CAD model.
From this model, multiple different views with known viewpoints are generated (fig.~\ref{fig:view_generation}) that represent the classes for comparison with the real object.

This way of modelling the object implicitly makes an important assumption: The object is centered in the view of the camera.
Since the views are taken from viewpoints that are located on a sphere (or multiple spheres of varying radii) with the same focus point, this model only accounts for 4 degrees of freedom, namely the distance to the object and yaw, pitch, roll of the camera.
This means that the remaining two dimensions (the unknown offset of the object with respect to the focus point) render the pose estimate ambiguous.
For objects located far enough from the camera, this does not directly pose an issue since the missing two parameters can be retrieved through an additional alignment step.
Here, 2D-based methods often rely on template matching \cite{Hinterstoisser2012} or learning-based solutions \cite{Xiang2017} in order to retrieve the location of the object, whereas 3D-based methods compute a local reference frame \cite{Aldoma2012} of the segmented scene object to achieve the same.

% \begin{figure}[!t]
%     % \vspace{1ex}
%     \centering
%     \includegraphics[width=0.45\columnwidth]{figures/view_cameras_scene.png}
%     \includegraphics[width=0.45\columnwidth]{figures/view_cameras_model.png}
%     \includegraphics[width=0.45\columnwidth]{figures/view_far_image_scene.png}
%     \includegraphics[width=0.45\columnwidth]{figures/view_far_image_model.png}
%     \includegraphics[width=0.45\columnwidth]{figures/view_close_image_scene.png}
%     \includegraphics[width=0.45\columnwidth]{figures/view_close_image_model.png}
%     % \includegraphics[height=175px]{figures/ambiguity_far.png}
%     % \includegraphics[width=0.45\columnwidth]{figures/ambiguity_model.png}
%     % \includegraphics[height=175px]{figures/ambiguity_close.png}
%     \caption{View on an engine from (left) the real camera and (right) the corresponding synthetic view  modelled with the focus point in the center of the object. The difference in camera pose can be computed using the offset. (middle row) At long distances, visible parts of the observed object match the scene object. (bottom row) At close distances, the real and generated view show huge differences in the visibility of the object's surface and the perspective distortion.}
%     \label{fig:ambiguity}
%     % \vspace{-5ex}
% \end{figure}
\begin{figure}[!t]
    \vspace{1ex}
    \centering
    \includegraphics[height=175px]{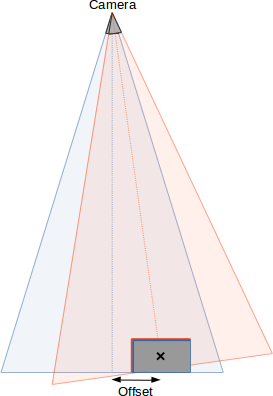}
    \includegraphics[height=175px]{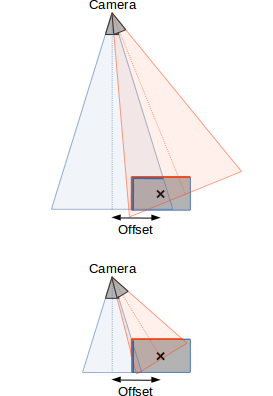}
    \caption{View on an engine from the real camera (blue) and the corresponding synthetic view (red) modelled with the focus point in the center of the object. The difference in camera pose can be computed using the offset. (left) At long distances, visible parts of the observed object match the scene object. (top right) When approaching the object, the real and synthetic view start to deviate. (bottom row) At close distances, the real and generated view show huge differences in the visibility of the object's surface.}
    \label{fig:ambiguity}
    \vspace{-3ex}
\end{figure}

\begin{figure}[!t]
    \vspace{1ex}
    \centering
    \fbox{\includegraphics[width=0.45\columnwidth]{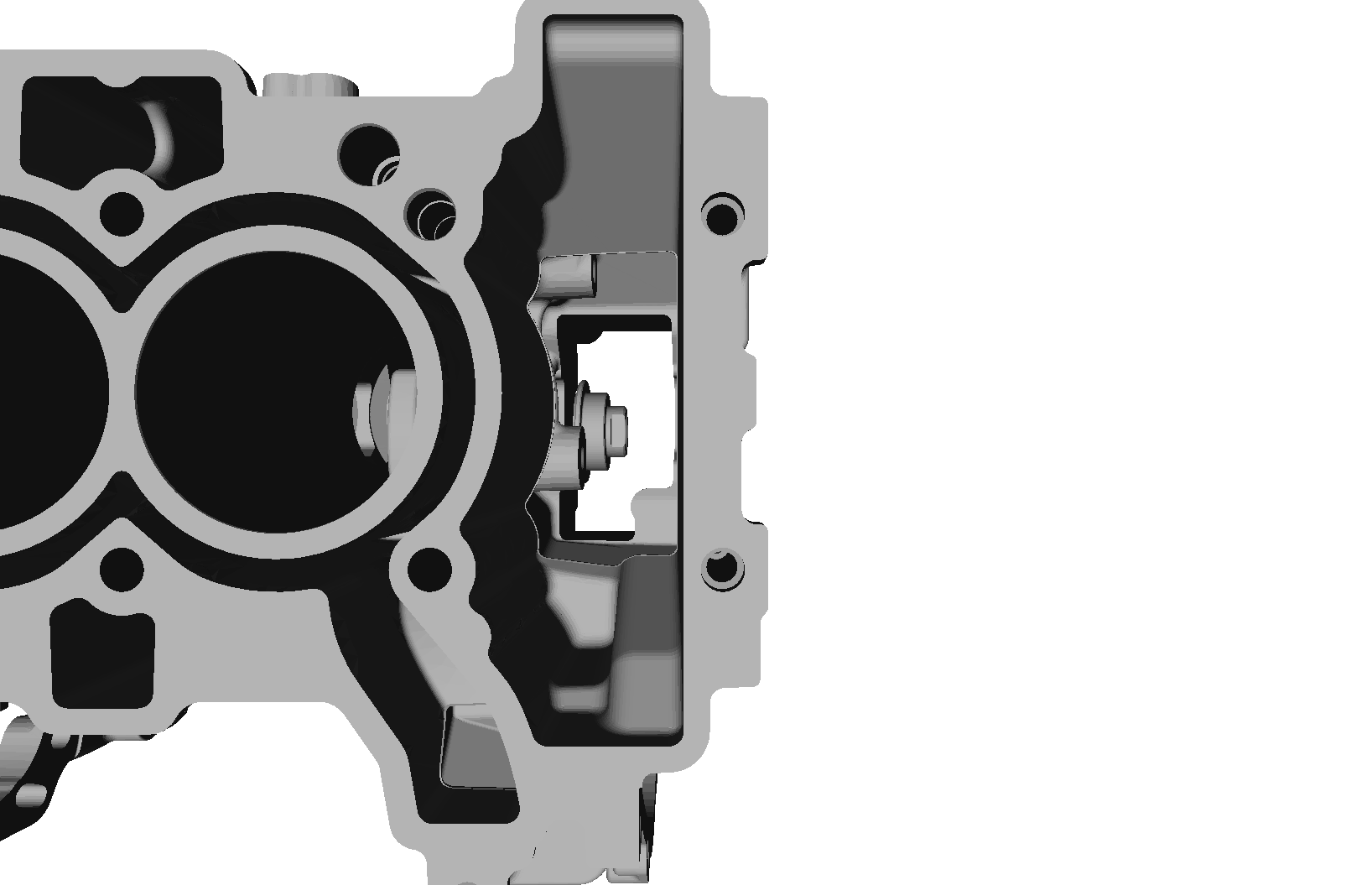}}
    \fbox{\includegraphics[width=0.45\columnwidth]{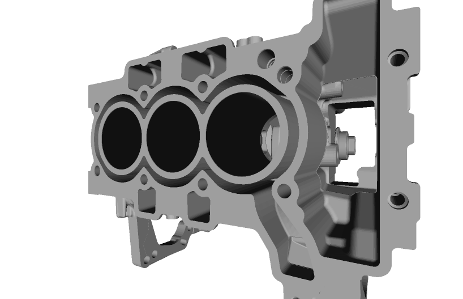}}
    \caption{An example of a close range view on an engine block: Difference between the real view (left) and the corresponding modelled view (right) show the shortcomings of typical view-based methods in close range: Different surface areas of the object are visible in addition to the perspective distortion.}
    \label{fig:ambiguity_image}
    \vspace{-3ex}
\end{figure}

However, the initial assumption of a centered object implies that the view generation uses a slightly different camera pose compared to the real scene.
This change of perspective can be usually ignored when the object is further away (depending on the focal length of the camera), since we can assume that the object remains fully in view and the projection of the object on the image plane can be considered almost orthographic (fig.~\ref{fig:ambiguity}, left).
When moving the camera towards the object, however, these assumptions do not hold anymore (fig.~\ref{fig:ambiguity}, right):
The object might not be fully in view anymore such that changes of the viewing direction of the camera will expose more or less visible parts of the object.
Identifying the correct view under these circumstances becomes a difficult challenge for any classifier, as the real and synthetic view differ too much.
Additionally, in close distance to the object, we have to deal with perspective distortion (traversal and axial magnification) of the object, which reduces the similarity between correct views even more (fig.~\ref{fig:ambiguity_image}).
% Especially for 2D-based approaches, this is particularly difficult to handle: Global methods suffer from the curse of dimensionality, while local feature-based methods require a highly realistic rendering of the object and it's environment to get good visual matches.

An obvious solution to this problem is to adapt the view generation to cover all 6 dimension, instead of modelling only 4 degrees of freedom.
However, this usually becomes computational intractable due to the curse of dimensionality, especially when we try to compute all possible views beforehand.
Instead, the generation of the synthetic views should be performed online only for relevant views.
However, it is not trivial to find these relevant views, since we still have to search the entire high-dimensional view space.
In order to find camera poses with a higher probability of generating views similar to the observed scene object, we have to know the object's pose in the first place.
This chicken-or-egg problem can however be solved recursively by implementing a smart search strategy and by tracking promising views over time, as described in section~\ref{sec:approach}.

%% file: approach.tex
\section{Sequential Bayesian pose estimation\\*with gradient-ascent}\label{sec:approach}
\noindent
% The solution to the close-range pose estimation problem has to consider the following requirements:
% First, the approach has to estimate all 6 parameters of the object's pose simultaneously to avoid differences of the real and generated views as discussed beforehand.
% Second, due to the higher dimensionality of the problem, we cannot rely on precomputed views, but have to generate them on-the-fly.
% Third, a method to evaluate the similarity between observed and generated object view has to be defined.
% Fourth, we need to deal with motions of the target object which might occur during the task execution.
% Fifth, the tracking requires a closed-loop system iterating between arm motion and pose estimation. It therefore has to work close to real-time to make it useful in practice.
In our previous work \cite{Grossmann2017}, we showed that the formulation of a sequential Bayes filter in form of a particle filter is a well-suited method to integrate multiple observations over time to resolve the ambiguities of single-shot pose estimation methods.
Here, we extend this approach to accommodate the algorithm for close-range observations and moving objects.

\subsection{Particle filter}
\noindent
The sequential Bayes filter is governed by two alternating steps of \emph{updating} (eq.~\eqref{eq:pose:filter:update}) the belief of the current state (pose estimate) given all observations $z_{0:t}$ and \emph{predicting} (eq.~\eqref{eq:pose:filter:predict}) the next state given the current state $\theta_t$, the current control input $u_t$ and all observations $z_{0:t}$.
\begin{align}
p(\theta_t | z_{0:t}) &\propto p(z_t | \theta_t) \ p(\theta_t | z_{0:t-1}) \label{eq:pose:filter:update} \\
p(\theta_{t+1} | z_{0:t}) &= \int p(\theta_{t+1} | \theta_{t}, z_{t}, u_{t+1}) \ p(\theta_{t} | z_{0:t} )d\theta_{t}\enspace. \label{eq:pose:filter:predict}
\end{align}
The control input $u_t$ is used to incorporate the directly measured motion of the robot arm's wrist (mounting point for the camera).
We model the state space in terms of \emph{full} viewing poses (which is equivalent to the inverse of the object's pose) given as a 6-dimensional vector for the translational and rotational part encoded in a transformation matrix:
\begin{align}
	\theta = (t_x, t_y, t_z, \alpha_{roll}, \beta_{pitch}, \gamma_{yaw})\,\widehat{=}\,[R\,|\,\vec{t}] \enspace,
\end{align}
whereas the observations $z$ are captured with a depth sensor in the form of a 3D point cloud.
The definition of a sequential Bayesian filter for representing pose probabilities requires the formulation of a \emph{motion model} $p(\theta_{t+1} | \theta_t, z_t, u_{t+1})$ and the \emph{observation model} $p(z_t | \theta_t)$.

% \todo[inline]{Can you repeat the difference with the other bayesian approaches?}

\subsection{Fast online view generation}

When considering the full 6-dof view space, the generation of synthetic views beforehand is not feasible anymore due to the high memory consumption and computational demands.
The view generation has therefore to be performed online and becomes a crucial factor for the time performance of the algorithm, since separate views have to be generated for the particles.

We implemented a fast method to generate point cloud views from a CAD file on-the-fly, as proposed in previous work \cite{Grossmann2017}.
Our approach works similar to the rasterization process of typical graphics pipelines, but instead of creating a projected image from the model, we are interested in the 3D points that are used to generate that image.
This could be done by reprojecting the image into 3D space again, however, this is computational more expensive and reduces the accuracy of the object's surface due to the discretization.
Therefore, we first sample the CAD model in a preprocessing step (only done once), and then filter non-visible points by projecting the points onto a depth buffer, keeping only the index of the closest point for each pixel.
Visible points can then be extracted using the indices.

\subsection{Observation model: Comparing views}
The observation model describes the likelihood of the observation given a viewing pose.
A direct comparison of a viewing pose (state) with a point cloud (observation) is not possible.
We therefore encode the viewing pose in a joint space with the observation.
This requires a partial model (in industrial use cases often given as a CAD model) from a given viewing pose, which can be generated by the fast view generation described before.
The similarity measure for two point clouds can then be derived as follows:

% Following the global pose estimation approaches, this means that the view space has to be discretized and rendered from each perspective.
% This is not written well. It needs a lot of guessing to understand...

% In order to compare the observation of the object in the scene with the virtual views, we need to represent each view with a feature descriptor.
% However, global feature descriptors are primarily designed to be used for object recognition tasks, as outlined in \ref{sec:background:global:descriptor}.
% It means that they are able to robustly discriminate between two very different views, but for similar views, especially in presence of missing data, they tend to lose their discriminative power and are not applicable for our industrial objects.
% Global descriptors that try to compensate for the sensitivity to missing data usually depend on clusters of smooth surfaces, and they are thus not suited to describe the objects either.
% Therefore, in our approach, we decided to not rely on feature descriptors at all and work directly with the point cloud representation of the object.
% Since we do not need to compute any local (e.g. surface normals) or global features, we can also reduce the computational cost immensely.
% \comment{Shorten paragraph above?, vok: I am fine with it. It feels redundant, but since it is becomes important below, it is ok.}
%
Given a point cloud of a rendered CAD model with $m$ points $p_m \in M$ from the viewing pose $\theta_t$, the observation point cloud $z_t$ with points $p_v \in O$ and a correspondence map $c_t \in C: M \mapsto O$ that maps $n$ model points to observation points $(p_v^k,p_m^k) \in c_t$, we define our similarity measure as the mean squared error between two point cloud:
\begin{align}
	f(c_t, \theta_t) = \frac{1}{n} \sum_{k=1}^n \| \R{\theta} \cdot p_v^k + \t{\theta} - p_m^k \|^2 \label{eq:MSPE}
\end{align}

\begin{figure}[!t]
    \vspace{1ex}
    \centering
    \includegraphics[height=100px]{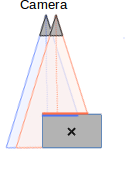}
    \scalebox{-1}[1]{\includegraphics[height=100px]{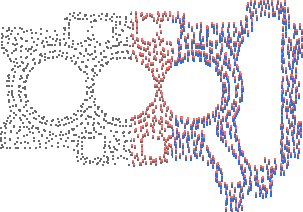}}
    \caption{Two parallel views of an object: the real view (blue) and a synthetic view (red) have almost the same point cloud. The only way to differentiate them is to account for the overlap ratio.}
    \label{fig:ambiguity_cloud}
    \vspace{-3ex}
\end{figure}

However, since we are modelling the full 6-dimensional view space, we also allow generated views where the object is only partially visible.
Two parallel views which only differ in the size of the visible area (fig.~\ref{fig:ambiguity_cloud}) are therefore not directly distinguishable with our similarity metric.
In order to compensate for this error, we scale the similarity with the overlap ratio between the real view and the synthetic views, based on the found correspondences,

However, the correspondences between real and synthetic view are usually unknown.
We apply a straight-forward scheme, which is also used in the \emph{iterative closest point} (ICP) algorithm, by simply selecting the nearest neighbors within a limited distance in the real view for each point in the synthetic view.
Additionally, we apply a \emph{sample consensus} (SAC) correspondence rejector to be more robust against outliers, e.g. due to sensor noise.
Integrating the overlap ratio ($n$ matches given $m$ model points), we can now define our log-likelihood of the observation given the current state and the correspondences as
\begin{align}
	\log p(z_t|\theta_t, c_t) = \frac{m}{n} \cdot f(c_t, \theta_t)
\end{align}

\subsection{Motion model: Gradient ascend}

In most sequential Bayesian filters, we assume that the proposal distribution is fully determined by the transition prior probability distribution
\begin{align}
	q(\theta_t | \theta_{0:t-1}, u_{0:t}, z_{0:t}) = \pi(\theta_t | \theta_{t-1}, u_t) \quad.
	\label{eq:proposal}
\end{align}
% where $z_t$ denotes the observation at time $t$ and $\tilde{\theta}_t$ is the predicted state.
The motion model then defines the transition from one state $\theta_t$ to the next predicted state $\theta_{t+1}$ and is defined by the recursive state transition function
\begin{align}
	& \theta_{t+1} = h(\theta_t, u_t) \enspace.
\end{align}

%In other words, the motion model reflects the search direction in the state space.
%In the simplest case, one would use a diffusion step here (Gaussian noise) that explores the state space evenly\footnote{Note that the state and noise are multiplied since we are dealing with poses (transformation matrices) here.}:
%\begin{align}
%    & \theta_{t+1} = \theta_t \cdot N(0,\Sigma) \enspace. \label{eq:pose:filter:motion:diffusion}
%\end{align}

% Additionally, a control input such as the sensor motion, if available, can be included to reflect knowledge about the state evolution.
% In our case, such control input is not required which makes the pose estimation also viable when the motion cannot be determined such as in hand-held sensors or with moving objects.

Due to the high-dimensional space of potential poses, we direct the search towards the more likely states based on the Langevin Monte Carlo method \cite{Nemeth2016} by using the gradient of the log-likelihood function.
With the gradient in the motion model, the state transition function reads
\begin{align}\label{eq:pose:filter:motion:gradient}
      \theta_{t+1} &= \theta_t \cdot u_t \cdot N(0,\Sigma) \cdot \gamma \nabla \log \theta_t \nonumber \\
\text{with} \quad \nabla \log \theta_t &= \nabla \log p(z_t|c_t,\theta_t) 
\end{align}
where $\gamma$ is the step-size of the gradient\footnote{For brevity, the scaling of the gradient is written as a simple multiplication with the scaling factor $\gamma$.
Since we are dealing with transformation matrices, this operation is actually a linear interpolation in SE(3).}.
Eq.~\ref{eq:pose:filter:motion:gradient} can also be seen as a gradient ascent method (neglecting the noise and control input, as they can be applied afterwards).
\begin{align}
   \hat{\theta}_{t+1} = \theta_t \cdot \gamma\nabla \log p(z_t|c_t, \theta_t) \enspace .
\end{align}
Intuitively, we are therefore trying to find the state $\hat{\theta}$ that maximizes the likelihood.
However, the gradient of the log-likelihood cannot be solved analytically, since it relies on the correspondence estimation itself.
The correct computation of the correspondences $\hat{c_t}$, in contrast, requires a good pose estimate $\hat{\theta}$ in the first place.
Instead of computing the gradient directly, we express this problem as a maximization-maximization problem \cite{Neal1998} with
\begin{align}\label{eq:em}
   \hat{c_t}    &= \arg \max_{c_t} p(z_t|c_t,\theta_t) \\
   \hat{\theta} &= \arg \max_{\theta} p(z_t|c_t,\theta_t)\enspace.
\end{align}
Since the observation model also requires the correspondences, we exploit this and use the very same ICP to solve this problem.
Here, we are only interested in the gradient and not the maximum, therefore a few iterations (2-3) are sufficient for the gradient estimation.
Note that the step-size $\gamma$ is implicitly retrieved during the estimation process. 

Using the gradient in the motion model has several advantages:
The particles are drawn to local maxima representing multiple modes of the posterior capturing multiple pose hypothesis at the same time.
We can therefore cover the posterior in the high-dimensional state space with less particles while converging faster to potential modes.
Even when the modes of the posterior shift rather rapidly due to large object movements, we are able to track the object's pose.
Since particles in areas of low probability will move towards higher probabilities, it also fights the problem of particle degeneracy and reduces the need for frequent resampling.
The particle diffusion, on the other hand, works as an antagonist for the gradient-ascent and avoids particle impoverishment, i.e. the particles located in a local maxima will not collapse to a single point.

\subsection{Implementation and Optimization}

Our implementation is based on a particle filter that has been extended with a gradient-ascent-driven motion model.
The input for our algorithm is given as CAD model (or dense point cloud) and a segmented object in form of a point cloud from the real scene.
Additionally, we use the measured arm motion as a control input to further optimize our motion model.
We initialize our particles on a sphere centered around the observed object with a radius equal to the mean distance.
Resampling has only to be done when the effective sample size drops below $50$\%.
The updating of the particles is done in parallel:
We first propagate and diffuse the particles, and then generate the according views, followed by 2-3 ICP iterations.
Finally, we compute the log-likelihood using the mean squared error between the observed and synthetic point cloud scaled by the overlap.
In order to optimize the execution time, we do all computations in the log-space.
As a result, the weight update becomes a simple sum, which we scale by $0.5$ to ensure that the weights stay in a representable range.
Unnormalized weights do not pose an issue, as we are only interested in the MAP estimate, which is scale independent.
Additionally, due to the log-space, the MAP estimate has now to be retrieved as the particle with the minimum weight.

% \vspace{-4mm}
\begin{algorithm}
    \KwData{Model $M$, Motions $U=u_{0:t}$, Observations $Z=z_{0:t}$}
    \KwData{Particles (Pose, Weight) $\langle P,W \rangle$}
    \KwResult{Pose estimate $\hat{\theta}$}
    $dist \gets \textit{mean}(z_0)$\\
    $P \gets \textit{sample}(\text{dist})$\\
    \ForEach{($z_k, u_k) \in (Z,U)$}{
        \If{$\textit{effectiveSampleSize} < 0.5$}{
            $P \gets \textit{resample}(P,W)$\\
        }
        \ForEach{$p_i \in P$}{
            $p_i \gets \textit{propagate}(p_i, u_k)$\\
            $p_i \gets \textit{diffuse}(p_i, \sigma^2)$\\
            $m_i \gets \textit{generateView}(p_i, M)$\\
            $t_i \gets p_i^{-1}$\\
            \For{0 \KwTo N}{
                $m_i \gets \textit{transform}(m_i, t_i)$\\
                $c_i \gets \textit{nearestNeighbors}(z_k, m_i)$\\
                $c_i \gets \textit{rejectCorrespondences}(c_i)$\\
                $t_i \gets \textit{leastSquares}(z_k^{[c_i]}, m_i^{[c_i]})$\\
            }
            $p_i \gets t_i^{-1}$\\
            $\epsilon_i \gets \frac{\textit{\#}m_i}{\textit{\#}c_i} \cdot \textit{MSE}(z_k^{[c_i]}, m_i^{[c_i]})$\\
            $w_i \gets 0.5*(w_i + \epsilon_i)$
        }
        $\textit{map} \gets \arg \min_i w_i$\\
        $\hat{\theta}_k \gets p_\textit{map}$\\
    }
    \caption{Particle filter with gradient-ascend}
    \label{alg:particle_filter}
\end{algorithm}

% The EM algorithm is then implemented as in alg.~\ref{alg:particle_filter} with 10 iteration of the SAC correspondence rejector and 4 ICP iterations.

%% file: evaluation.tex
\section{Application and Evaluation}

%The proposed pose estimation approach has been applied on two realistic industrial use-cases in the context of a engine assembly: a piston insertion and a rod cap placing.
%In both cases, the arm approaches the engine from above where only a small part of the engine is visible to the camera.
%While moving the arm to a roughly known position above the engine, we are estimating the engine's pose in order to insert the piston / crankcase cap in the correct place.
%The major challenge is to compute an accurate estimate without ever having the full object in view during the trajectory execution.

% \begin{figure*}[!h]
%     \centering
%     \includegraphics[width=\textwidth]{figures/rodcap_seq1.jpg}
%     \caption{Typical sequence of a crankcase cap placement ($dt=1 \textit{sec}$). During the sequence, the engine block is never fully in view.} 
%     \label{fig:rodcap:seq1}
% \end{figure*}

\begin{figure*}[ht]
    \centering
    \includegraphics[width=\textwidth]{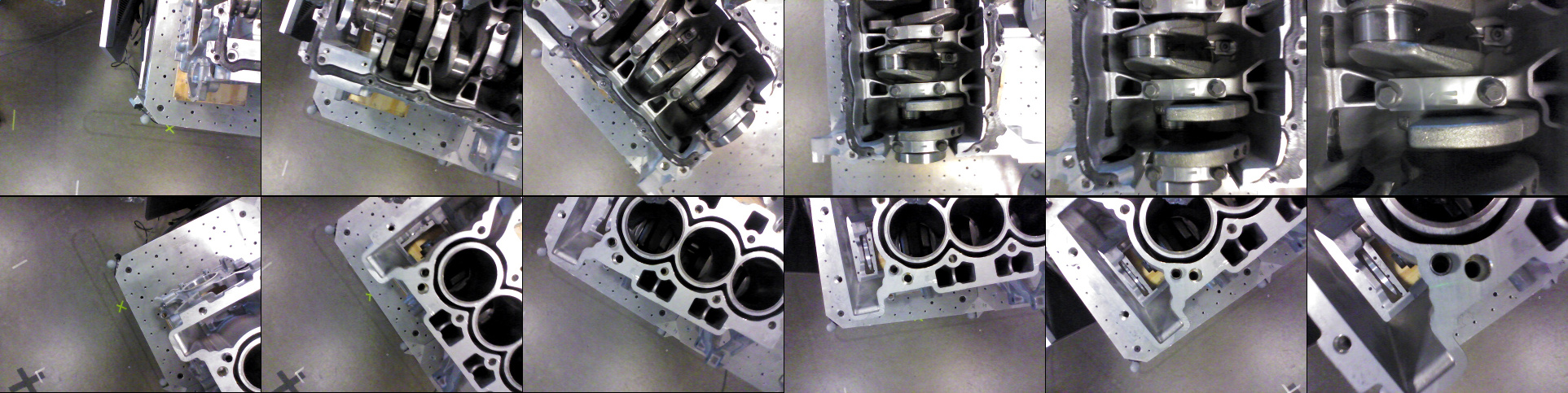}
    \caption{Typical sequence of (top) a piston insertion and (bottom) a crankcase cap placement. During the sequence, the engine block is never fully in view.} 
    \label{fig:piston:seq1}
    \vspace{-4mm}
\end{figure*}

For our experiments, we use a \textit{kuka iiwa} robot arm with a \textit{RealSense F200} wrist-mounted camera.
A typical sequence of captured images during the tasks is shown in Fig.~\ref{fig:piston:seq1}.
%The pose estimation algorithm is running on a \textit{Intel Xeon} CPU (2.4GHz) with 8 cores and 12 GB of RAM without any GPU optimizations, however, the particle filter is parallelized.
The ground truth for the engine's pose has been generated semi-automatically by aligning a fully visible, discriminative view to the model, which we afterwards verified manually.
Furthermore, we recorded the trajectory of the camera during the process to backtrack the engine's pose over time.
The particle filter runs with only 200 particles in all experiments.
In order to evaluate our algorithm, we ran multiple trials.
For each trial, we used 1 baseline test (see Sect. \ref{sec:baseline}, below) (with the full view of the engine and a static camera) and 4 realistic task executions from different starting poses for the piston insertion and crankcase cap placement respectively.
For each trial, we ran multiple alignments, which in all cases converged to the correct baseline pose.
% As outlined above, a thorough quantitative introduction is difficult so that we take also a practical approach during the evaluation and 
%However, since the pose estimation is heavily dependent on the starting poses and partial views observed used during the process, the approach is only hard to quantify and cannot be displayed in a condensed form.
%Therefore, we will 
% demonstrate the performance of the algorithm by the means of concrete examples and outline the characteristics and limitation of the approach.

\begin{figure}[!ht]
    \vspace{1ex}
    \centering
    \includegraphics[width=0.9\columnwidth]{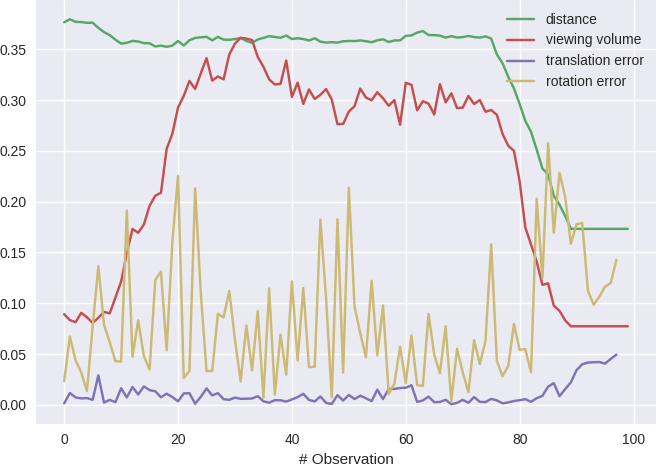}
    \caption{Evaluation of the pose estimation error with respect to the distance and viewing volume of the camera.} 
    \label{fig:partial}
    \vspace{-1em}
\end{figure}

\subsection{Baseline}\label{sec:baseline}
For the baseline, we conducted a simple experiment with a static camera placed roughly $40 cm$ over the engine block. From that view, the engine was fully visible. The approach used for the baseline is the one from \cite{Grossmann2017} which was shown (and since then has proven) to be precise.
We evaluated the pose estimation 5 times to compute the precision and accuracy of the algorithm in terms of the mean error and the standard deviation.
With the whole object in view and neglecting the burn-in phase of the particle filter, we derive a mean error of $3.2mm \pm 1.3mm$ for the translational error and $3.3\deg \pm 2.2\deg$ after convergence. Here, the error is computed with respect to the object's coordinate frame (center). Note that the accuracy of the points on the visible surface is usually even higher. We can easily handle this magnitude of errors during assembly using the compliance of our IIWA Kuka arms \cite{Rovida2018}.

\subsection{Close-range partial views}
The goal of this experiment was to determine the impact of the camera distance to the object, and thereby the ratio of the visible parts of the object, on the outcome of the pose estimate.
Therefore, we captured the distance of the object based on the mean of the point cloud and its extend by computing an ellipsoid around the visible points using its principle axes.
The volume of the ellipsoid gives a rough estimate of the visibility ratio.

When setting the distance and volume in contrast with the error development (here: error norm), as shown in fig.~\ref{fig:partial}, we can see that especially when the camera reduces the distance to the object in the end of the sequence (cf. fig.~\ref{fig:piston:seq1}), the viewing volume reduces to a point where the view becomes too ambiguous and the particle filter starts failing at a distance less than $25cm$ (around observation 85). Depending on the used sensor (due to increasing noise in short distances), this value might differ and has to be empirically estimated for other RGB-D cameras.

\subsection{Piston insertion and crankcase cap placement}

\begin{figure*}[!ht]
    \centering
    \includegraphics[width=0.9\columnwidth]{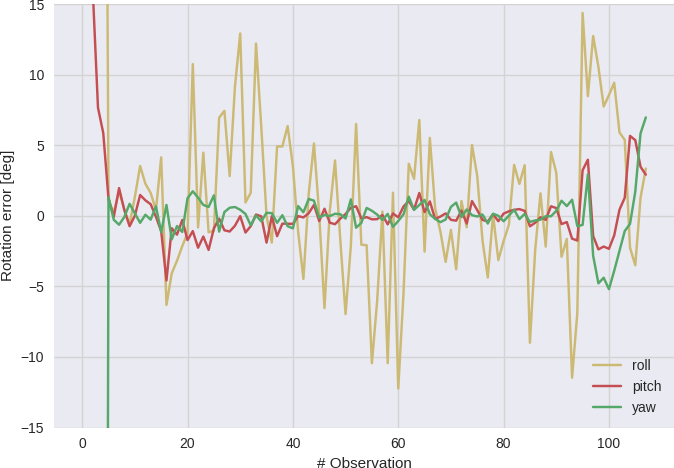}\quad
    \includegraphics[width=0.9\columnwidth]{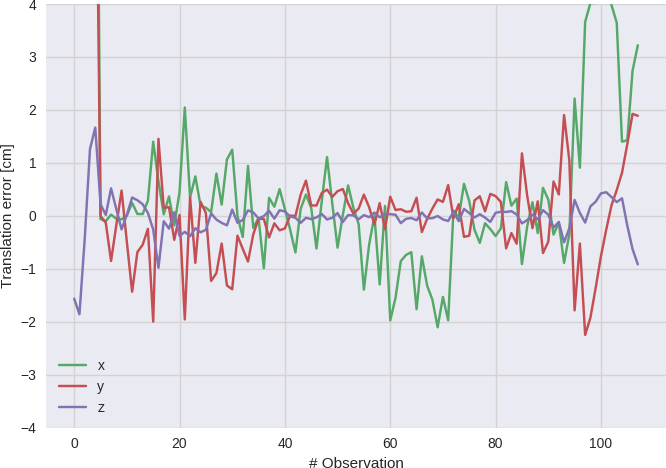}
    \\~\\
    \includegraphics[width=0.9\columnwidth]{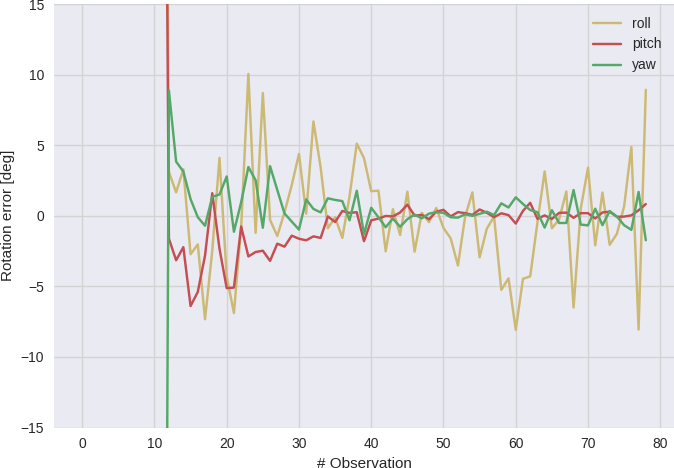}\quad
    \includegraphics[width=0.9\columnwidth]{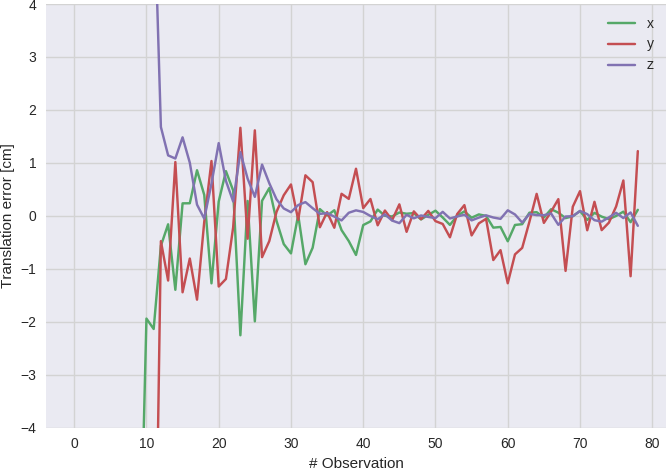}
\caption{(top) Piston insertion and (bottom) Crankcase cap placement: Rotational and translational errors while approaching the top of the engine block from a random direction.} 
    \label{fig:eval}
    \vspace{-4mm}
\end{figure*}

The main experiments were conducted under a realistic movement of the robot arm during (fig.~\ref{fig:eval}, top row) the piston insertion and (fig.~\ref{fig:eval}, bottom row) the crankcase cap placement.
As usual, the particle filter requires a burn-in phase of 10 to 15 observations.
Afterwards, the particle filter is, in both cases, able to converge rather quickly to the real pose.
The estimated pose during the piston insertion converges with an average translational errors (x,y,z) and maximal deviations of $(6.1\pm5.5, 3.9\pm3.6, 0.8\pm0.8)mm$ and a rotational error and maximal deviation of $(0.5\pm0.4, 0.6\pm0.5, 3.9\pm3.2)\deg$.
% what is the \pm?!? std-deviations? max deviation?
%
The pose during the crankcase cap placing converges slightly slower to the correct pose with an average translational error (x,y,z) and maximal deviations of $(0.8\pm0.9, 3.1\pm2.9, 0.5\pm0.4)mm$ and a rotational error and maximal deviation of $(0.4\pm0.4, 0.2\pm0.2, 2.2\pm1.9)\deg$.
The higher accuracy here can be explained by the structure of the engine block: compared to the piston insertion, where the motor is viewed from the top with little amount of structure, the bottom view looks into the motor and has much more depth information available for the alignment.

In both cases, the error in depth is the smallest translational error.
This is related to the alignment of the points using ICP, since displacements in depth (without nearby corresponding points) are recovered immediately.
Movements along the object's surface are not so easily detected and depend mainly on a discriminative surface structure.
The majority of the rotational error is related to the camera roll.
This is a general problem, but it is amplified when using partial views, since the visible surface of the object is in most cases rectangular.
Registration algorithms, such as the ICP, therefore prefer to align the rectangular structure than the often noisy surface structure of the object.

%% file: conclusions.tex
\section{Conclusions}

In this paper, we presented a fast particle-based pose estimation method using multiple observations over time.
We focused on scenarios where robots with wrist-mounted cameras continuously have to estimate and track an objects pose while approaching and interacting with it.
In these cases, observations can only be obtained from a close range and we have to deal with very limited partial views of the objects, which poses a challenge to view-based pose estimation algorithms.
We show that these methods are making the assumption of a 4-dimensional view space for generating synthetic model views, which leads to wrong perspectives in close distance.
We proposed to solve this issue by modelling the full 6D view space and rendering the views of the model online.
Our experiments demonstrate in a realistic assembly task that, by only using partial close-range observations, our algorithm is able to compute the pose of the work piece with a reasonably high accuracy, accurate enough to deal with many manufacturing tasks.
In cases where we need higher accuracy, the compliance of a robot arm and force-based exploration strategies can be exploited to cope with the residual uncertainty.

In the introduction, we emphasized that cycle time is critical and that the pose estimation approach must be fast.
Particle filter in higher dimension are known to be rather slow, but our approach based on the gradient search strategy, allows us to run the algorithm in almost real-time (1.5Hz) with plenty of space for optimizations.
We reach this speed on a \textit{Intel Xeon} CPU (2.4GHz) with 8 cores and 12 GB of RAM.
The particle filter is parallelized using all 8 cores. GPU optimization is not used.

Even though our approach gives good initial results and shows that the concept holds, it also has its limitations:
One of the major drawback is that even though we can cope with a certain amount of noise due to the probabilistic approach, we still require the target object to be segmented from the scene, which by itself is a non-trivial task.
However, since our approach models the full pose of the camera (or object), it actually allows the free positioning of virtual cameras and with an improved observation model, it could be used directly in the scene to estimate the pose of one or multiple objects.

Furthermore, at a certain distance, even after convergence of the particle filter, the prior is not strong enough to keep the correct pose estimate anymore.
In the experiments we found empirically that for the Intel RealSense camera, a threshold of $25$cm was a good cut-off point to stop capturing more images and to accept the computed pose.
However, this threshold is use-case dependent and would have to be adapted for other depth sensors and objects.
We will explore this cut-off threshold more systematically in the future.